\documentclass{fairmeta}

\usepackage[utf8]{inputenc}
\usepackage{amsmath}
\usepackage{amssymb}
\usepackage{amsfonts}
\usepackage{nicefrac}
\usepackage{enumitem}
\usepackage{url}

\title{Gaze2Act: Gaze-Conditioned Vision-Language-Action Policies for Interactive Robot Manipulation}
\renewcommand{\titlelist}{{\LARGE\sffamily\bfseries
\begin{minipage}{0.96\linewidth}
\setstretch{1.08}
\raggedright
Gaze2Act: Gaze-Conditioned Vision-Language-Action\\
Policies for Interactive Robot Manipulation
\end{minipage}}}

\author[1,*]{Kuangji Zuo}
\author[1,*]{Gen Li}
\author[1]{Bofan Lyu}
\author[1]{Yanshuo Lu}
\author[1]{Boyu Ma}
\author[1]{Shijia Han}
\author[1]{Xinyu Zhou}
\author[1]{Xichen Yuan}
\author[1]{Chuhao Zhou}
\author[1]{Jiaqi Bai}
\author[1]{Geng Li}
\author[1,\dagger]{Jianfei Yang}

\affiliation[1]{MARS Lab, Nanyang Technological University}

\contribution[*]{Equal contribution}
\contribution[\dagger]{Corresponding author}

\abstract{Vision-Language-Action (VLA) models have recently shown strong potential for robot learning by following language instructions.
However, in practice, language alone is often insufficient to precisely convey human intent.
It is difficult to describe which exact object to interact with among similar candidates, where to act on the object, or how the target may change during execution.
To address this limitation, we propose Gaze2Act, a novel VLA framework that leverages human gaze as a dynamic and intuitive intent signal for complex interactive manipulation.
Gaze2Act first bridges the ego-exo view gap by mapping first-person gaze into the robot's perspective through cross-view semantic matching, producing both an object mask and a gaze point for coarse-to-fine target specification.
These cues are then integrated into the policy through perception-level prompting and action-level conditioning, allowing the robot to attend to relevant regions and execute precise interactions under dynamic intent.
In a systematic evaluation across seven task categories and 16 real-robot tasks on a Unitree G1 humanoid, Gaze2Act achieves state-of-the-art performance in both intent accuracy and task success rate.
It notably outperforms baselines in object disambiguation, fine-grained interaction, and dynamic intent steering.
These results demonstrate that human gaze provides a natural, low-burden, and highly expressive modality for human-in-the-loop VLA control.
}

\correspondence{Jianfei Yang: \email{jianfei.yang@ntu.edu.sg}; Kuangji Zuo: \email{kuangji001@e.ntu.edu.sg}}
\metadata[Project]{\url{https://zuo-kuangji.github.io/Gaze2Act/}\par\vspace*{0.32cm}}

\begin{document}

\maketitle
\begin{figure}[t]
\centering
\includegraphics[width=\textwidth]{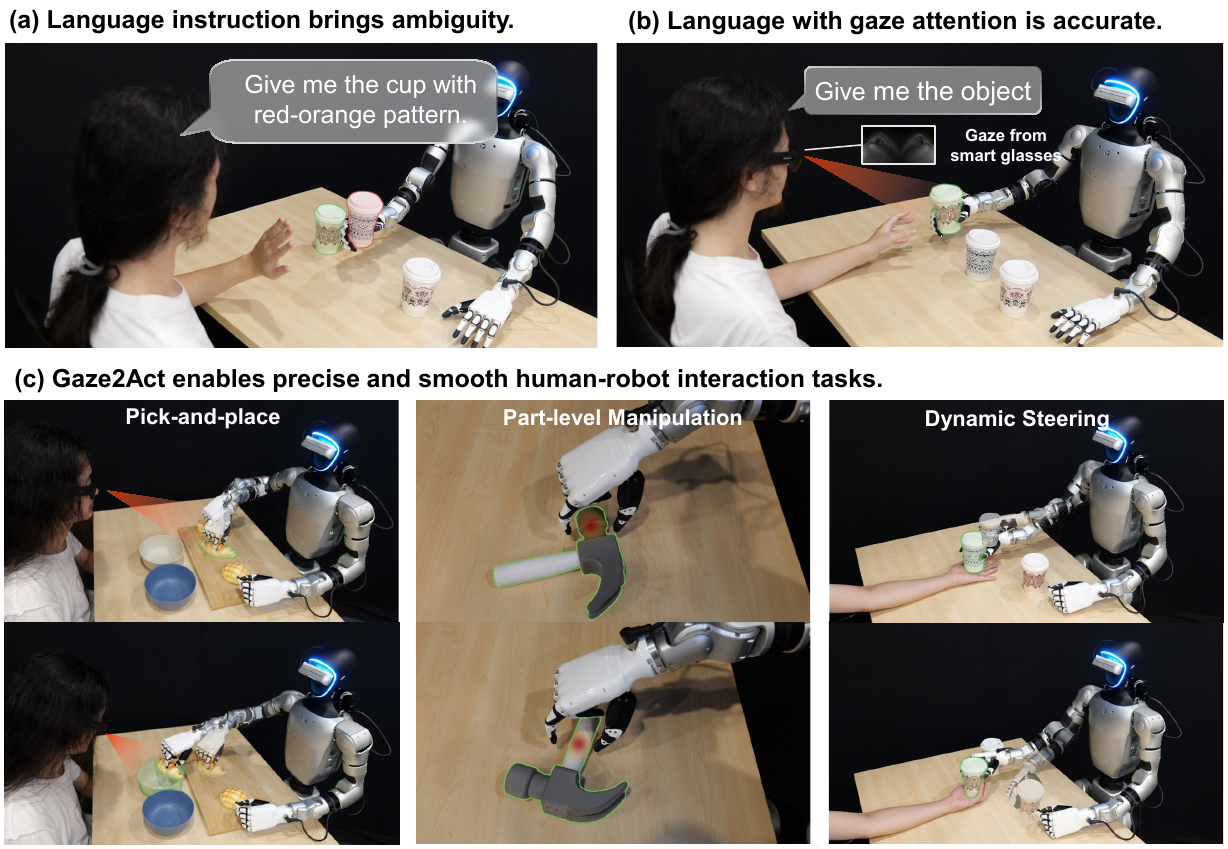}
\caption{Motivation of gaze in human-robot interaction and new capabilities enabled by Gaze2Act.}
\label{fig:overview}
\end{figure}

\section{Introduction}
\label{sec:intro}

Robotic systems are increasingly expected to follow human intent in complex, real-world interactions.
However, conveying intent precisely remains a fundamental challenge.
In practice, users rarely specify their goals in a fully explicit and unambiguous manner: they may need to select one object among visually identical candidates, indicate a fine-grained interaction location~\citep{mo2021where2act}, or revise their target dynamically during execution.
Such intent is inherently spatial, fine-grained, and time-varying, making it difficult to express through static and discrete interfaces. 

Vision-Language-Action (VLA) models~\citep{black2024pi0, bjorck2025gr00t, kim2024openvla, liu2026palm, lin2025evo0,lin2025evo1,lin2026evo} represent a promising step toward bridging perception and action through a unified language interface.
However, most existing VLA systems rely on language as the primary channel for specifying intent, implicitly assuming that intent can be fully described in words.
This creates a fundamental bottleneck: language alone often lacks the precision and temporal continuity required for interactive, real-time robot manipulation.

Recent work attempts to mitigate this limitation by translating language into explicit spatial representations, such as bounding boxes or segmentation masks, generated by vision-language models or language-prompted detectors~\citep{affgrasp, ellmer, huang2025roboground, li2025controlvla, yu2025point, wang2026vp}.
However, these approaches still inherit the intrinsic ambiguity of language.
Linguists have found that over 20\% of expressions are descriptive, i.e., implicit referring expressions~\citep{levinson1983pragmatics, krahmer2012computational}, which makes it difficult for robots to identify the correct object~\citep{jiang2026reibench}.
As a result, these VLA models struggle to disambiguate between similar objects and cannot naturally support dynamic intent updates during execution.
The core limitation, therefore, is not the lack of spatial representations but the absence of a direct and continuously updated intent signal.
This highlights the need for a complementary, non-linguistic interface that can convey spatial intent more precisely and adaptively.

Human gaze provides such an interface.
Decades of cognitive studies show that gaze precedes and guides manipulation, with the eyes fixating on the target before the hand begins to move~\citep{johansson2001eyehand, flanagan2003action}.
This eye-leads-hand coordination indicates that gaze is not merely observational, but a direct manifestation of motor intention.
Unlike language, gaze is continuous, immediate, and spatially precise: it naturally disambiguates targets, specifies interaction locations, and reflects intent shifts in real time, as shown in Fig.~\ref{fig:overview}.
Beyond its behavioral significance, gaze is also becoming increasingly accessible as a practical interaction modality.
With eye tracking rapidly integrated into AR/VR and XR glasses from major industry players, gaze is emerging as a new paradigm for human-machine interaction, with growing real-world adoption.
Taken together, these properties make gaze not only a complementary signal, but a fundamentally different and potentially indispensable channel for conveying human intent in embodied systems.
This raises a key question: \textit{how can human gaze be seamlessly and effectively integrated into VLA-based robot manipulation?}

To answer this question, we propose Gaze2Act, a framework that augments VLA policies with human gaze.
Gaze2Act uses language to specify the high-level task objective, while leveraging gaze to resolve target ambiguity and provide fine-grained spatial cues during execution.
A central challenge is that human gaze is observed from a first-person perspective and cannot be directly aligned with the robot’s viewpoint.
We address this by introducing a marker-free cross-view semantic matching mechanism based on visual foundation models, which projects gaze into the robot’s observation space without requiring camera calibration, task-specific training, or external markers.
Once aligned, Gaze2Act integrates gaze-conditioned intent into the policy through perception-level prompting and action-level grounding. This allows the robot to attend to relevant regions and execute precise, dynamically updated interactions, while preserving the prior knowledge of the pretrained policy. 
Our contributions are summarized as follows:
\begin{itemize}[leftmargin=*]
    \item We propose \textbf{human gaze as a continuous and dynamic intent interface for VLA policies}, enabling precise target specification across object-level, part-level, and dynamically evolving interaction scenarios.
    \item We propose \textbf{Gaze2Act, a framework that grounds first-person gaze into the robot’s visual space}, producing target masks and interaction points through marker-free cross-view alignment, and integrating these cues via perception-level visual prompting and action-level grounding.
    \item \textbf{A comprehensive real-world evaluation on a Unitree G1 humanoid, covering seven task categories and 16 tasks}, demonstrates consistent improvements in object disambiguation, fine-grained localization, multi-target specification, and dynamic intent steering.
\end{itemize}

\section{Related Work}
\label{sec:related}

\subsection{Gaze-Guided Robot Manipulation}
\label{sec:related_gaze}

Gaze is a non-invasive behavioral cue that closely reflects human intent and has received sustained attention in human-robot interaction and robot manipulation research. Prior work mainly exploits gaze from two perspectives. The first treats gaze as a signal for offline behavior modeling and intent learning. Gaze-VLM~\citep{pani2025gazevlm} uses gaze to assist VLMs in egocentric activity understanding covering targets, actions, and intent, while GazeVLA~\citep{li2026gazevla} further treats gaze as a supervision signal for intention learning, improving the model's ability to capture human manipulation intent. 
The second uses gaze to assist robot manipulation online. GAMMA~\citep{tay2026gamma} combines real-time gaze from AR glasses with VLM reasoning and uses foundation models for cross-view target association to invoke parameterized skills, and \citet{shafti2019gaze} use gaze to infer targets for assisted grasping. 
Overall, these studies demonstrate the value of gaze for manipulation understanding, target selection, and robot assistance. However, gaze is typically used as offline supervision or within modular pipelines rather than as a direct, continuous control signal. Gaze2Act, by contrast, incorporates gaze-driven guidance in both training and inference of a VLA policy, enabling continuous integration of user intent during execution.

\subsection{Explicit Spatial Conditioning for VLA Policies}
\label{sec:related_spatial}

VLA policies typically take language instructions as the primary task condition, relying on the model to implicitly infer spatial information such as target locations, manipulation regions, and placement areas. Recent work augments VLA grounding through explicit spatial conditioning. One line of methods passes language-derived target regions as intermediate spatial representations to the policy. For example, RoboGround~\citep{huang2025roboground} leverages object and placement masks from language and visual observations as spatial guidance, while ControlVLA~\citep{li2025controlvla} injects object-centric conditions into action generation via ControlNet-style conditioning~\citep{zhang2023controlnet}. Another line overlays spatial anchors directly onto visual observation. In particular, VP-VLA~\citep{wang2026vp} and Point-VLA~\citep{yu2025point} employ structured visual prompts, such as crosshairs, points, and bounding boxes, to explicitly specify manipulation targets and regions. These methods demonstrate that spatial conditioning can substantially improve target localization and manipulation performance. However, the spatial signals come from language parsing rather than the user's intent during execution. When language is imprecise or intent shifts during execution, such signals may fail to capture user preference. Gaze2Act, by contrast, derives object- and part-level conditions directly from human gaze, enabling continuous incorporation of user intent during execution.

\section{Method}
\label{sec:method}

\begin{figure}[t]
\centering
\includegraphics[width=\textwidth]{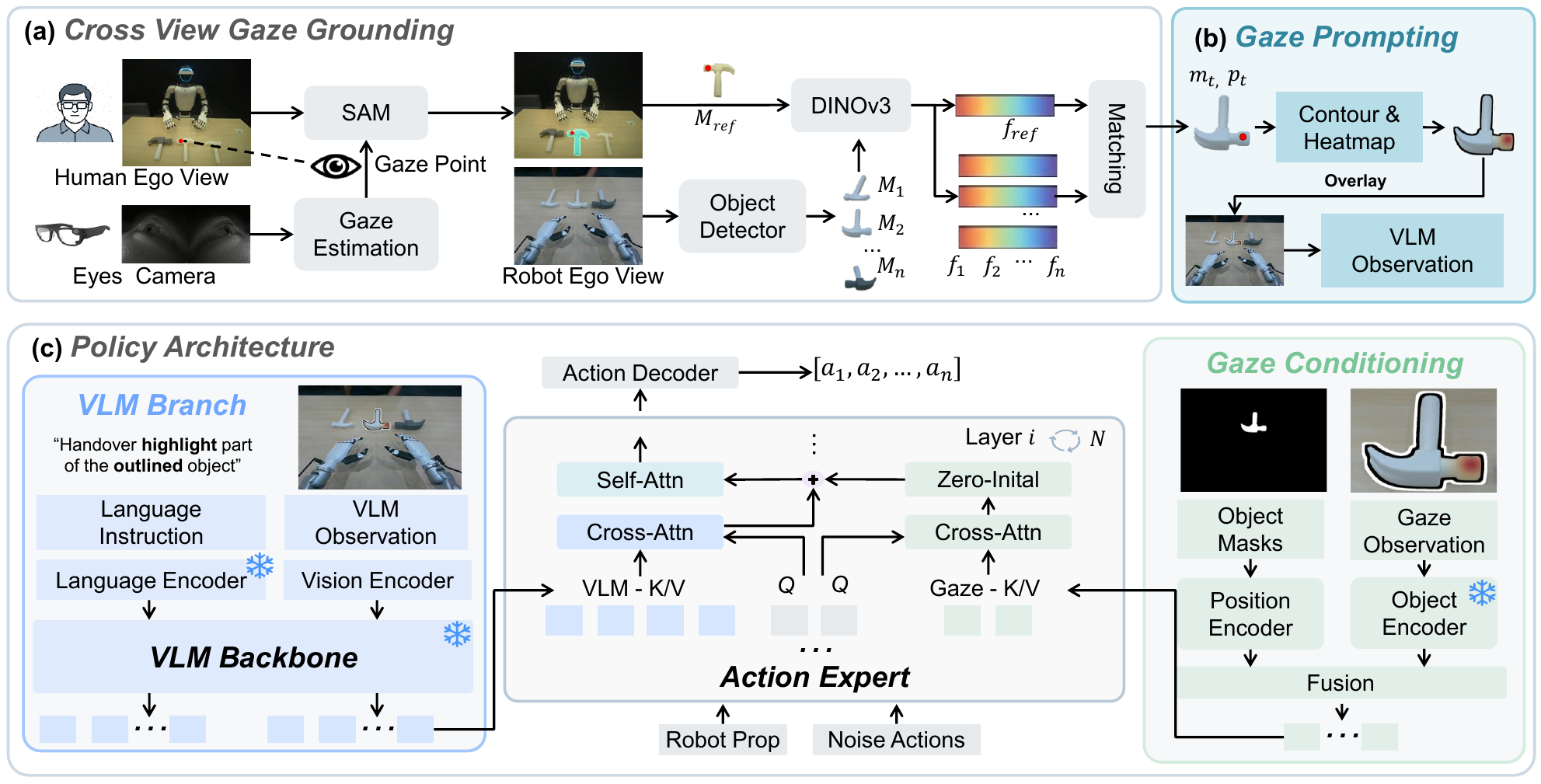}
\caption{\textbf{Overview of Gaze2Act.} \textbf{(a) Cross-view gaze grounding} aligns the operator's first-person gaze with the robot view and produces gaze-grounded target cues. \textbf{(b) Perception-level gaze prompting} renders the grounded gaze cues as visual prompts on the robot observation, including object contours and gaze heatmaps. \textbf{(c) Policy architecture with action-level gaze conditioning} integrates gaze-conditioned intent into the Vision-Language-Action (VLA) policy. The Vision-Language Model (VLM) branch encodes the language instruction and gaze-prompted observation, while the gaze-conditioning branch injects gaze-grounded spatial tokens into the Diffusion Transformer (DiT) action expert through a zero-initialized decoupled cross-attention branch.}

\label{fig:method}
\end{figure}

\subsection{Overview}
\label{sec:overview}

A standard VLA policy $\pi_\theta$ maps a language instruction $l$ and visual observation $o_t$ to an action sequence $a_t = \pi_\theta(l, o_t, s_t)$, where $s_t$ denotes the proprioceptive state. Gaze2Act introduces a first-person gaze coordinate $g_t \in \mathbb{R}^2$ (provided by AR/XR glasses) as an additional conditioning signal, enabling the target referent to be continuously updated by the user’s gaze during execution. We implement Gaze2Act on top of GROOT N1.5~\citep{bjorck2025gr00t}, a representative VLA backbone, in which a frozen Eagle2 VLM~\citep{li2025eagle} encodes $(l, o_t, s_t)$ into multimodal tokens $\mathbf{z} \in \mathbb{R}^{N \times d}$ that condition a DiT~\citep{peebles2023scalable} action head through cross-attention. As illustrated in Fig.~\ref{fig:method}, gaze conditioning proceeds in three stages. Cross-view gaze grounding (Section~\ref{sec:grounding}) maps $g_t$ to a target mask $m_t$ and fixation point $p_t$ in the robot view. Perception-level gaze prompting (Section~\ref{sec:vp}) renders $(m_t, p_t)$ onto the observation as visual prompts: object contours give coarse-grained target specification, and gaze-centered heatmaps produce fine-grained interaction cues. Action-level gaze conditioning (Section~\ref{sec:gcb}) injects the gaze token into the denoising action head through decoupled cross-attention.

\subsection{Cross-View Gaze Grounding}
\label{sec:grounding}

To ground the first-person gaze coordinates $g_t$ in the robot observation $o_t$, we propose a multi-granularity matching framework that is training-free and calibration-free, as illustrated in Fig.~\ref{fig:method} (a). The matching process operates sequentially across coarse and fine granularities to identify the fixated target mask $m_t$ and fixation point $p_t$, respectively. 
We adopt DINOv3~\citep{simeoni2025dinov3} ViT-L/16 and extract features from five uniformly sampled transformer layers ($\{0, 5, 10, 15, 20\}$).
The patch tokens from these layers are then L2-normalized and averaged to obtain the final feature map $\phi(\cdot)\in\mathbb{R}^{H\times W\times D}$.
This multi-layer aggregation preserves both low-level texture and high-level semantic cues, improving robustness under viewpoint and appearance variations. In deployment, gaze grounding is voice-triggered, and the selected target is tracked until the next trigger, removing the need for continuous fixation on the intended target during interaction.

\noindent\textbf{Coarse-grained matching.} Given $g_t$ as a point prompt, we first employ SAM3~\citep{carion2025sam3} to segment the reference object from the egocentric image $o_t^{\text{ego}}$, producing a reference mask $M^{\text{ref}}$. SAM3 is then applied in automatic mode on the robot observation $o_t$ to generate a set of candidate masks $\{M_k\}_{k=1}^K$. The patch features within each mask are averaged to obtain $\mathbf{f}_{\text{ref}}$ and $\mathbf{f}_k$, and the mask with the highest cosine similarity to the reference feature is then selected:
\begin{equation}
\mathbf{f}_{\text{ref}}=\frac{1}{|M^{\text{ref}}|}\sum_{p\in M^{\text{ref}}}\phi(o_t^{\text{ego}})[p],\quad
\mathbf{f}_k=\frac{1}{|M_k|}\sum_{p\in M_k}\phi(o_t)[p],\quad
m_t=\arg\max_{M_k}\cos(\mathbf{f}_{\text{ref}},\mathbf{f}_k).
\label{eq:obj_match}
\end{equation}

\noindent\textbf{Fine-grained matching.} To localize the target part within the selected object, we perform a second-stage matching conditioned on the coarse mask $m_t$.
Since the point prompt $g_t$ provides only a sparse indication of the target part location, we extract a $(2R{+}1)\times(2R{+}1)$ neighborhood around its corresponding patch on $o_t^{\text{ego}}$ and average the features to obtain a robust reference vector $\mathbf{f}_{\text{ref}}^{\text{fine}}$. This vector is compared against the dense features of $o_t$ via cosine similarity to identify the corresponding point, with the argmax restricted to the coarse-grained mask $m_t$ obtained in the previous step:
\begin{equation}
p_t = \underset{(i,j)\in m_t}{\arg\max}\;
\cos\bigl(\mathbf{f}_{\text{ref}}^{\text{fine}},\,\phi(o_t)[i,j]\bigr).
\label{eq:part_match}
\end{equation}
This constraint prevents the part-level match from being distracted by visually similar objects elsewhere in the scene. 
Together the two stages output $(m_t, p_t)$, which serve as a unified spatial representation used in the perception layer (Section~\ref{sec:vp}) and the action layer (Section~\ref{sec:gcb}) of the policy. 

\subsection{Perception-Level Gaze Prompting}
\label{sec:vp}

Given the target representation $(m_t, p_t)$ from Section~\ref{sec:grounding}, we first overlay gaze-derived cues onto the robot observation at the perception level, as illustrated in Fig.~\ref{fig:method} (b). Object contours provide coarse-grained target specification, while gaze-centered heatmaps provide fine-grained interaction cues. These signals are directly rendered on the robot observation image, allowing the policy to receive spatial guidance through the visual channel.

\noindent\textbf{Contour overlay.} To provide object-level grounding, a colored contour is drawn along the boundary of $m_t$ on $o_t$ to produce $o_t'$, with different colors distinguishing different targets in multi-target tasks. The language instructions can be category-agnostic (e.g., ``pick up the outlined object''), with object identity fully grounded in the visual modality rather than the language input.

\noindent\textbf{Gaze heatmap.} 
While $p_t$ provides a precise localization signal, directly using a single point is brittle for visual policy learning. We therefore convert it into a continuous spatial heatmap to provide a soft and robust guidance signal.
Specifically, a two-dimensional isotropic Gaussian heatmap $\mathcal{H}_t$ is superimposed within the $m_t$ region, centered at $p_t$:
\begin{equation}
\mathcal{H}_t(x, y) = \exp\!\left(-\frac{(x - p_t^x)^2 + (y - p_t^y)^2}{2\sigma^2}\right) \cdot \mathbf{1}[(x, y) \in m_t],
\label{eq:heatmap}
\end{equation}
where $\sigma$ is the scale hyperparameter of the Gaussian kernel. 
During manipulation, gaze provides fine-grained spatial guidance only in the pre-contact phase. 
The two types of visual prompts are selected based on the spatial granularity required by the task: fine-grained, part-level interactions rely on heatmap prompting, whereas coarse object-level actions are sufficiently guided by contour cues alone. Appendix~\ref{app:gaze_prompt_design} provides further details on this design.

\subsection{Action-Level Gaze Conditioning}
\label{sec:gcb}

Perception-level gaze prompting makes the gaze-selected region visible to the VLM, but using gaze only as an image overlay leaves the spatial condition implicit. We therefore introduce action-level gaze conditioning that encodes the grounded mask and fixation point as a compact spatial token and injects it directly into the DiT action head, as described in Fig.~\ref{fig:method} (c). This structured conditioning complements the visual prompt by explicitly incorporating gaze into the denoising process.

\noindent\textbf{Gaze token construction.} The representation $(m_t, p_t)$ is decoupled into position and object pathways and encoded separately. The position pathway uses a two-dimensional sinusoidal code, computed at $p_t$ when $p_t \in m_t$ and at the center of $m_t$ otherwise, yielding $\mathbf{z}_t^{\text{pos}} \in \mathbb{R}^{d_p}$. The object pathway uses a frozen 29M-parameter DINOv3 ViT-S+/16~\citep{simeoni2025dinov3} as the object encoder. It encodes the target crop from the bounding box of $m_t$, resized to $224\times 224$, and outputs the CLS feature $\mathbf{z}_t^{\text{obj}} \in \mathbb{R}^{d_v}$. The two pathways are concatenated and projected to the backbone dimension through a linear layer $W_{\text{fuse}}$, forming a single gaze-grounded spatial token (gaze token) $\mathbf{c}_t^{\text{gaze}} \in \mathbb{R}^{d}$:
\begin{equation}
\mathbf{c}_t^{\text{gaze}} = W_{\text{fuse}}\,[\mathbf{z}_t^{\text{pos}};\, \mathbf{z}_t^{\text{obj}}] + b_{\text{fuse}}.
\label{eq:gaze_token}
\end{equation}

\noindent\textbf{Decoupled cross-attention.} Within each DiT block, Gaze2Act adds an independent cross-attention path in parallel with the original vision-language path~\citep{ye2023ipadapter}. The two paths share the same action query but use separate key-value projections, allowing the original path to preserve global vision-language context while the new path injects gaze-derived spatial constraints. Self-attention and the feed-forward network (FFN) remain unchanged from the pretrained backbone. Both branches share the same Query $\mathbf{Q} = W_Q\,\mathbf{h}$, where $\mathbf{h}$ is the noisy action and robot proprioception hidden representation. Their Keys and Values, however, come from separate sources: the original path projects VLM tokens $\mathbf{z}$ through pre-trained weights $W_K, W_V$ to obtain $(\mathbf{K}, \mathbf{V})$, while the new path projects the gaze token $\mathbf{c}^{\text{gaze}}$ through newly added weights $W_K^{\text{gaze}}, W_V^{\text{gaze}}$ to obtain $(\mathbf{K}^{\text{gaze}}, \mathbf{V}^{\text{gaze}})$. The outputs of both paths are summed as a residual into the cross-attention sub-module output:
\begin{equation}
\mathbf{h}_{\text{xattn}} = \mathbf{h} + \underbrace{\mathrm{Attn}(\mathbf{Q}, \mathbf{K}, \mathbf{V})}_{\text{vision-language condition}} + \underbrace{\mathrm{Attn}(\mathbf{Q}, \mathbf{K}^{\text{gaze}}, \mathbf{V}^{\text{gaze}})}_{\text{gaze spatial condition}}.
\label{eq:dual_attn}
\end{equation}
The resulting $\mathbf{h}_{\text{xattn}}$ then passes through the original FFN sub-module to the next layer. The original path handles global language-vision context, and the new path handles gaze-derived spatial constraints, both acting at every denoising step and every layer.

\noindent\textbf{Stabilized gaze injection.} To preserve the pretrained action prior at the start of fine-tuning, we zero-initialize the output projection of the gaze cross-attention branch while using standard initialization for the other projections. With this design, the added gaze branch is initially a no-op to the backbone output, and its influence is learned gradually from the manipulation objective. This design incurs only 4.95\% additional parameters; see Appendix~\ref{app:training_details_subsec}.

\section{Experiments}
\label{sec:exp}

We evaluate Gaze2Act on a Unitree G1 humanoid robot across seven categories and 16 real-robot tasks. The experiments are designed to validate three key capabilities of Gaze2Act:
\begin{itemize}
\item resolving object-level and compositional ambiguity under distractors through coarse-grained gaze.
\item improving part-level interaction by specifying where to act through fine-grained gaze.
\item updating the target referent during execution.
\end{itemize}
\subsection{Experimental Design}
\label{sec:setup}

\textbf{Datasets and Tasks.} We design 15 tasks for experiment evaluation, covering object selection, compositional target specification, part-level interaction, and dynamic intent steering. We construct training demonstrations for each task group, covering multiple scene layouts and instruction variants. The same trajectories are used for all methods, but converted into method-specific supervision. For Gaze2Act, target masks and fixation points are annotated offline and rendered as gaze-grounded visual prompts during training, so eye-tracking hardware is not required for collecting demonstrations. At inference time, gaze is obtained from Meta Aria glasses~\citep{ariaglasses} and grounded online into the robot view. By contrast, language-conditioned baselines use the most specific, unambiguous instruction available in each scene. Details of all tasks are presented in Appendix~\ref{app:tasks}. 

\textbf{Baselines.}~We compare against Vanilla GROOT, RoboGround~\citep{huang2025roboground}, and ControlVLA~\citep{li2025controlvla}. RoboGround uses GLaMM~\citep{GLaMM_2024_CVPR} to obtain language-grounded masks and injects them through a mask-guided perceiver, while ControlVLA uses Grounding DINO~\citep{liu2024groundingdino} and SAM2~\citep{ravi2025sam2} to segment language-specified masks and injects them through ControlNet-style fine-tuning. For fairness, all methods are implemented on the same GROOT N1.5 backbone. We report Intent Accuracy and Task Success; the robot platform and gaze interface are described in Appendix~\ref{app:robot_platform}, baseline implementations in Appendix~\ref{app:baseline_impl}, training details in Appendix~\ref{app:training_details_subsec}, and training demonstrations in Appendix~\ref{app:data}.

\subsection{Main Results}
\label{sec:main_results}

\begin{table}[t]
\caption{Main results across 15 manipulation tasks. Int.\ = Intent Accuracy; Suc.\ = Task Success. Results are percentages over 50 trials per task. ``--'' indicates that part-level intent accuracy is not applicable. Object-level Avg.\ is computed over the first 10 tasks, Part-level Avg.\ over the last 5 tasks, and Overall Avg.\ Suc.\ over all 15 tasks. Overall Avg.\ Int.\ is averaged over all applicable intent-accuracy entries; for baselines without part-specific conditioning, it is computed over the first 10 tasks only.}

\label{tab:main}
\centering
\resizebox{\textwidth}{!}{
\begin{tabular}{ll cc cc cc cc}
\toprule
 & & \multicolumn{2}{c}{Vanilla GROOT} & \multicolumn{2}{c}{RoboGround} & \multicolumn{2}{c}{ControlVLA} & \multicolumn{2}{c}{Gaze2Act (Ours)} \\
\cmidrule(lr){3-4} \cmidrule(lr){5-6} \cmidrule(lr){7-8} \cmidrule(lr){9-10}
Category & Task & Int. & Suc. & Int. & Suc. & Int. & Suc. & Int. & Suc. \\
\midrule
\multirow{3}{*}{Ambiguous Obj.} & Cup & 44 & 44 & 96 & 94 & 82 & 62 & 92 & 92 \\
 & Bread & 28 & 18 & 36 & 32 & 40 & 32 & 98 & 96 \\
 & Fruits & 30 & 20 & 66 & 56 & 80 & 44 & 100 & 94 \\
\midrule
\multirow{3}{*}{Unseen Obj.} & Cup & 44 & 36 & 84 & 74 & 76 & 62 & 90 & 88 \\
 & Bread & 38 & 16 & 44 & 28 & 82 & 54 & 96 & 86 \\
 & Fruits & 48 & 26 & 76 & 40 & 90 & 54 & 94 & 86 \\
\midrule
\multirow{2}{*}{Transparent Obj.} & Cup & 30 & 24 & 56 & 32 & 64 & 42 & 88 & 86 \\
 & Bottle & 20 & 14 & 32 & 24 & 40 & 28 & 88 & 84 \\
\midrule
\multirow{2}{*}{Compositional} & Pick bread place bowl & 30 & 26 & 38 & 34 & 42 & 34 & 96 & 94 \\
 & Pick paper ball place bin & 24 & 18 & 78 & 32 & 84 & 52 & 88 & 84 \\
\midrule
\multirow{3}{*}{Subpart Grasp} & Hammer (handle) & -- & 24 & -- & 26 & -- & 28 & 80 & 62 \\
 & Hammer (head) & -- & 18 & -- & 22 & -- & 24 & 76 & 64 \\
 & Hammer (neck) & -- & 22 & -- & 26 & -- & 24 & 70 & 68 \\
\midrule
\multirow{2}{*}{Part-cond. Act.} & Cup (handover) & -- & 22 & -- & 38 & -- & 42 & 90 & 88 \\
 & Cup (pour) & -- & 20 & -- & 36 & -- & 40 & 86 & 80 \\
\midrule
\multicolumn{2}{l}{Object-level Avg.} & 33.6 & 24.2 & 60.6 & 44.6 & 68.0 & 46.4 & \textbf{93.0} & \textbf{89.0} \\
\multicolumn{2}{l}{Part-level Avg.}   & --   & 21.2 & --   & 29.6 & --   & 31.6 & \textbf{80.4} & \textbf{72.4} \\
\multicolumn{2}{l}{Overall Avg.}      & 33.6 & 23.2 & 60.6 & 39.6 & 68.0 & 41.5 & \textbf{88.8} & \textbf{83.5} \\
\bottomrule
\end{tabular}
}
\end{table}

\begin{figure}[t]
\centering
\includegraphics[width=\textwidth]{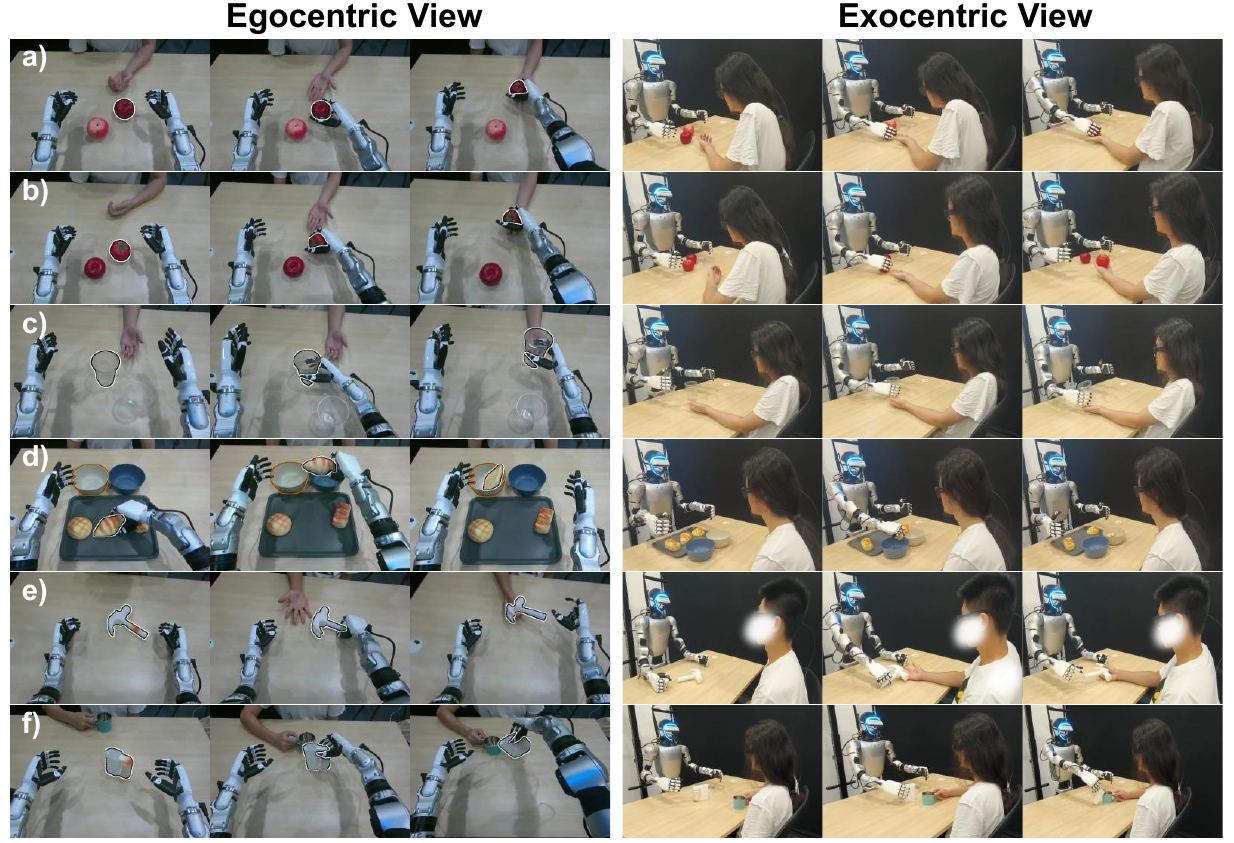}
\caption{Experimental demonstrations across representative tasks. Each row shows one task instance, with the human egocentric view on the left and the corresponding exocentric robot execution on the right. The rows correspond to (a) Ambiguous Obj., (b) Unseen Obj., (c) Transparent Obj., (d) Compositional, (e) Subpart Grasp, and (f) Part-cond. Act. The robot follows coarse- and fine-grained grounded gaze cues to complete different tasks.}
\label{fig:experiments}
\end{figure}

\begin{figure}[t]
\centering
\includegraphics[width=\textwidth]{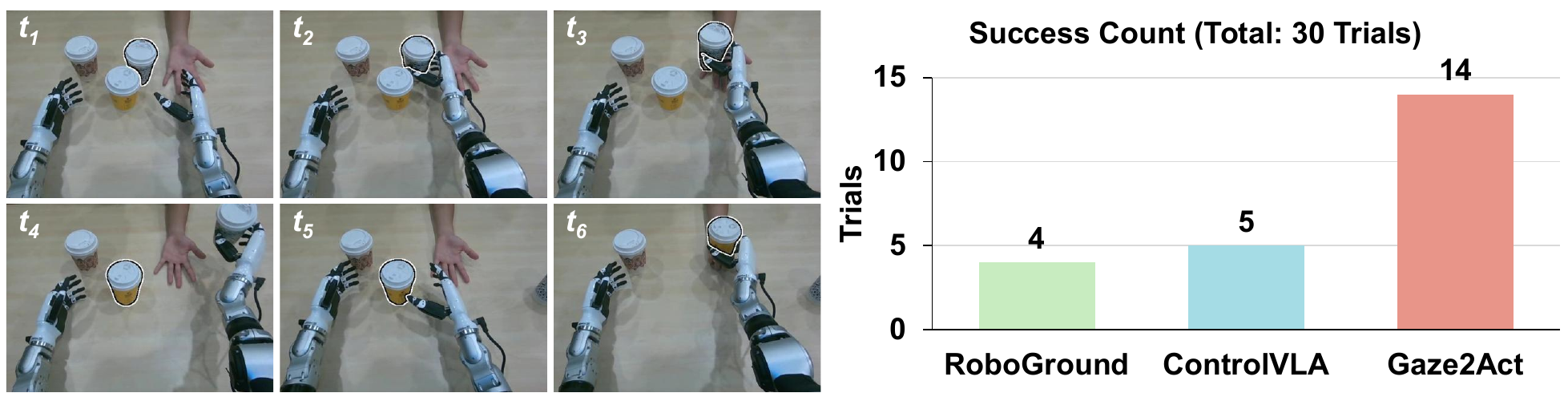}
\caption{Dynamic intent steering. \textbf{Left}: a real-robot rollout ($t_1$--$t_6$); the robot first moves toward one cup, the target changes during execution, and the robot redirects the action to hand over the newly intended cup. \textbf{Right}: success counts of the three methods over 30 target-switch trials.}
\label{fig:dynamic}
\end{figure}

Table~\ref{tab:main} answers the first two questions on static tasks. Object Intent and Compositional Intent test coarse-grained target specification under distractors, while Part Intent tests fine-grained interaction cues. Fig.~\ref{fig:dynamic} answers the third question by evaluating Dynamic Intent Steering with an unchanged language instruction.

\textbf{Object Intent.} In Ambiguous Obj., Unseen Obj., and Transparent Obj. tasks, the policy must identify one instance rather than a category. Vanilla GROOT remains unreliable even with specific attribute descriptions, reaching 34.0\% intent accuracy on Ambiguous Obj. tasks. RoboGround and ControlVLA improve over language alone when their masks are reliable, but the transparent rows show that detector based grounding can still degrade under difficult visual conditions. Gaze2Act is higher on every object selection row because the referent is supplied by gaze rather than inferred only from language attributes.

\textbf{Compositional Intent.} Compositional tasks require two referents: the object to operate and the placement target. Gaze2Act reaches 96\% and 88\% intent accuracy on the two compositional tasks, with 94\% and 84\% task success. The gap to the baselines is larger in success than in intent, which suggests that selecting plausible individual regions is not sufficient. The policy must also preserve the intended object and placement association throughout execution.

\textbf{Part Intent.} Task categories of Subpart Grasp and Part-cond. Act. answer the second question by testing fine-grained interaction cues. RoboGround and ControlVLA can complete these tasks with object masks, but they do not provide part-specific conditions, so part-level Int.\ is not applicable. Gaze2Act achieves 80.4\% part-level average intent accuracy and 72.4\% part-level average success. This supports the role of gaze in specifying where on the object the robot should interact. Hammer neck remains the hardest case because the target region is small and close to other graspable parts.

\textbf{Dynamic Intent Steering.} We evaluate target changes during execution in a long-horizon setting, where the policy must revise its intended target after the action has already started. This setting is challenging, with all methods achieving less than half success. RoboGround and ControlVLA can update their language-conditioned mask generators after the switch, but the updated mask signal is not sufficiently salient to redirect ongoing manipulation, achieving 4/30 and 5/30 success, respectively. Gaze2Act reaches 14/30 trials, remaining below half success, but performs substantially better because the updated gaze target is injected through both perception-level prompting and action-level conditioning. As shown in the rollout on the left of Fig.~\ref{fig:dynamic}, after the target changes, the policy correctly follows the newly grounded cup and completes the handover.

\subsection{Ablation Studies}
\label{sec:ablation}

\textbf{Ablation Setup.} We ablate the policy design on Pick Bread Place Bowl and Hammer (handle/head/neck), which together separate the two roles of gaze. Pick Bread Place Bowl tests whether gaze selects and binds the correct object and placement target. Hammer tests whether gaze preserves the intended contact point. Each task is evaluated over 60 trials. Hammer uses 20 trials per part across the handle, head, and neck.

\begin{table}[t]
\caption{Ablation on Policy Design Choices. ``Gaze prompting only'' uses a contour overlay for Pick Bread Place Bowl and a contour combined with a gaze heatmap for Hammer.}
\label{tab:ablation}
\centering
\begin{tabular}{lcc}
\toprule
Variant & Pick Bread Place Bowl & Hammer (handle/head/neck) \\
\midrule
Baseline                              & 17/60 & 15/60 \\
Gaze prompting only                 & 34/60 & 28/60 \\
Gaze conditioning only (random init)        & 24/60 & 17/60 \\
Gaze conditioning only (zero init)                     & 40/60 & 19/60 \\
Gaze2Act (full)                       & \textbf{55/60} & \textbf{39/60} \\
\bottomrule
\end{tabular}
\end{table}

\textbf{Complementary Gaze Pathways.} Table~\ref{tab:ablation} isolates three design choices: perception-level gaze prompting, action-level gaze conditioning, and stabilized injection. The ablation reveals a clear division of roles. For Hammer, visual prompting is the stronger single pathway, increasing the number of successful trials from 15 to 28, while action-level conditioning alone reaches 19 with zero initialization. This matches the nature of part-level manipulation, where the policy benefits from an explicit visual cue about where on the object the hand should interact. For Pick Bread Place Bowl, action-level conditioning is more effective, reaching 40 successful trials compared with 34 from visual prompting alone. This suggests that the gaze token helps the action head maintain the selected object and placement target during execution. The full model performs best on both tasks, with 55 successful trials on Pick Bread Place Bowl and 39 on Hammer, showing that the two pathways provide complementary information. Finally, random initialization weakens action-level conditioning, especially on Pick Bread Place Bowl, where performance drops from 40 to 24 successful trials. This supports the use of stabilized injection when adding a new gaze path to a pretrained action model.

\begin{figure}[t]
\centering
\includegraphics[width=\textwidth]{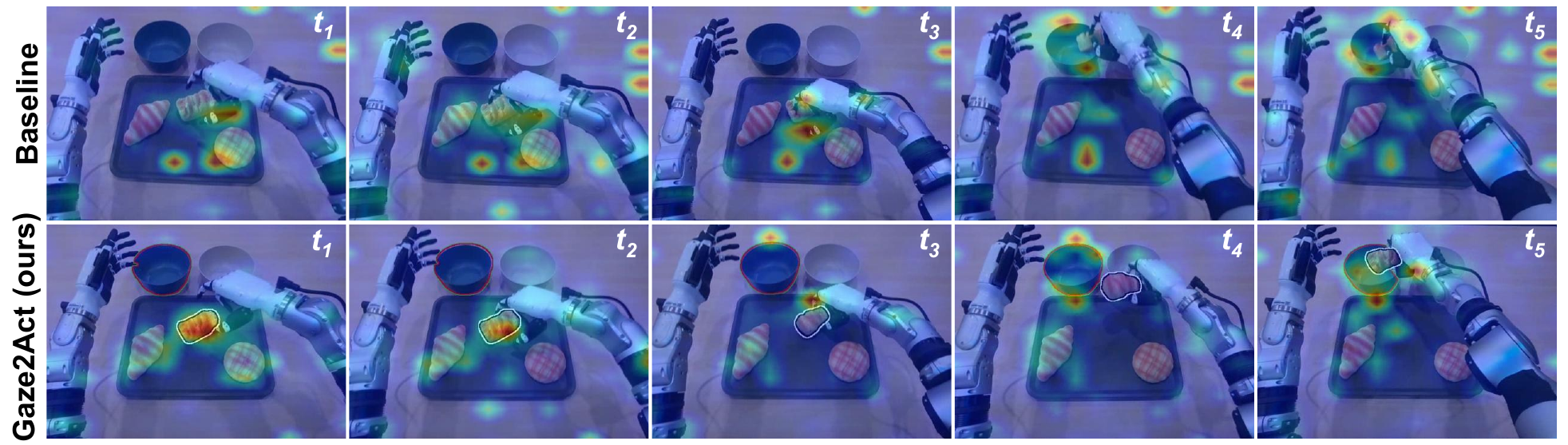}
\caption{Attention visualization on a Pick Bread Place Bowl task. Top row shows Vanilla GROOT with ``pick cream bun place white bowl.'' Bottom row shows Gaze2Act with ``pick outlined bread place outlined bowl.''}
\label{fig:attention}
\end{figure}

\textbf{Attention Analysis.}
\label{sec:attn_vis}
Fig.~\ref{fig:attention} shows the attention visualization of a Pick Bread Place Bowl task, where both the bread and the bowl have distractors. Vanilla GROOT receives the specific instruction ``pick cream bun, place white bowl,'' but its attention remains dispersed across the scene, with the grasp often landing near object edges and becoming less stable. Gaze2Act instead uses the generic instruction ``pick outlined bread place outlined bowl'' and obtains both referents from gaze. Perception-level gaze prompting marks the selected regions in the observation, while action-level gaze conditioning keeps these spatial cues explicit during denoising. Together, they produce a more concentrated spatial focus and a more stable grasp and place trajectory.

\section{Conclusion}
\label{sec:conclusion}
This paper presented Gaze2Act, a VLA framework that uses human gaze as an intent signal at inference for robot manipulation. Gaze2Act grounds first-person gaze in the robot view and uses the resulting mask and fixation point through perception-level gaze prompting and action-level gaze conditioning. This allows language to describe the task while gaze specifies the spatial referent. By combining visual prompts with direct action-level conditioning, Gaze2Act keeps gaze intent explicit from perception to action generation. Real-robot experiments on a Unitree G1 show that Gaze2Act improves coarse-grained target specification, fine-grained interaction, and Dynamic Intent Steering over language-only control and language-derived mask conditioning.

\section{Limitations}
Despite the promising results, Gaze2Act still has several limitations. First, the framework depends on reliable gaze estimation and cross-view grounding, which may become unstable under severe occlusion, rapid head motion, or large viewpoint discrepancies between the human and robot. Second, the current system assumes that gaze reflects the user’s intended manipulation target, whereas in practice human gaze may occasionally drift or exhibit exploratory behavior unrelated to the desired action. Future work will explore more robust gaze grounding, multimodal intent fusion, and stronger online policy adaptation mechanisms.
\bibliographystyle{assets/plainnat}
\bibliography{paper}

\clearpage
\appendix
\section{Experimental Details}
\label{app:exp_details}

\subsection{Robot Platform and Gaze Interface}
\label{app:robot_platform}
All real-robot experiments are conducted on a Unitree G1 humanoid robot. The policy observes the scene through a RealSense D455 RGB camera mounted on the robot. At inference time, the operator wears Meta Aria glasses, and first-person gaze coordinates are obtained with an open-source gaze estimation model. Gaze2Act follows an event-triggered interaction protocol. A voice keyword activates target selection; the current gaze point is used as an interactive point prompt to segment the intended target in the egocentric view, and the selected target is then tracked over time by SAM3. Therefore, the system does not re-run gaze-based target selection at every frame. A new voice keyword triggers a new gaze-based selection and cross-view matching step, which enables Dynamic Intent Steering.

\subsection{Baseline Implementations}
\label{app:baseline_impl}
All four methods use the same GROOT N1.5 backbone and are trained under the same computing environment and training schedule (Section~\ref{app:training_details_subsec}). They are trained from the same robot demonstration trajectories, but use different language conditioning, spatial supervision, and preprocessing pipelines according to each method.

\paragraph{Vanilla GROOT.} The original GROOT N1.5 policy without any explicit spatial conditioning. The Eagle2 VLM consumes the language instruction directly, and the DiT action head receives only the language-vision tokens through standard cross-attention.

\paragraph{RoboGround.} We follow Huang et al.~\citep{huang2025roboground} and use the publicly released GLaMM~\citep{GLaMM_2024_CVPR} checkpoint as the mask generator. The masks are concatenated channel-wise with the observation and processed through a Grounded Perceiver, a token resampler whose attention is guided by the masks, before being passed to the DiT action head. During training, the policy receives ground-truth object masks from data annotation. At the inference stage, GLaMM takes the robot observation and a specific language instruction as input, generates the target masks once at episode onset, and the masks are reused throughout execution.

\paragraph{ControlVLA.} We follow Li et al.~\citep{li2025controlvla} and use Grounding DINO~\citep{liu2024groundingdino} and SAM2~\citep{ravi2025sam2} as the mask generator. The masks are encoded into object-centric features and injected into the DiT action head via ControlNet-style~\citep{zhang2023controlnet} fine-tuning. During training, the policy receives ground-truth object masks from data annotation. At inference, Grounding DINO and SAM2 take the robot observation and a specific language instruction as input to generate the target masks.

\paragraph{Backbone unification.} The original RoboGround uses a GR-1 policy backbone and the original ControlVLA uses a Diffusion Policy~\citep{chi2025diffusion} backbone. To enable a fair comparison with Gaze2Act, we re-implement both methods on the GROOT N1.5 backbone, preserving each method's mask generation pipeline and integration mechanism. All training hyperparameters and the computing environment are unified with Gaze2Act.

\paragraph{Language description protocol.} For language-conditioned baselines (RoboGround and ControlVLA), we provide the most specific unambiguous language description available in each scene, e.g., ``handover the red patterned cup with white lid.'' Gaze2Act uses category-generic templates, e.g., ``handover the outlined object,'' so that object identity, contact point, and dynamic target updates must be supplied by gaze.

\subsection{Gaze Prompt Design}
\label{app:gaze_prompt_design}

This section explains how the visual prompt form is chosen for different task categories. The choice follows the spatial granularity required by the task. Object-level and compositional tasks require coarse-grained target specification, while subpart and part-conditioned tasks require fine-grained interaction cues.

\noindent\textbf{Object-level prompts.}
For Ambiguous Obj., Unseen Obj., Transparent Obj., and Compositional tasks, the intent is to select the correct object or placement target among distractors. In these settings, the contour prompt is sufficient because it marks the boundary of the gaze-selected referent while leaving the object appearance visible. We do not add a heatmap because these tasks do not require a within-object interaction location. Adding a local point cue could unnecessarily bias the policy toward an arbitrary region on the selected object.

\noindent\textbf{Part-level prompts.}
For Subpart Grasp and Part-cond. Act., the intent is not only which object to manipulate, but also where on the object the robot should interact. We therefore combine the contour prompt with a gaze-centered heatmap. The contour provides object-level grounding, and the heatmap provides the fine-grained interaction cue within the selected object. This matches the task design in which different fixations on the same object correspond to different grasp regions or manipulation modes.

\noindent\textbf{Pre-contact fine-grained guidance.}
During manipulation, fine-grained gaze cues are most useful before contact, when the fixation still indicates the intended interaction region. We therefore render the heatmap only when the grounded fixation remains inside the current object mask, i.e., $p_t \in m_t$. When this condition is not satisfied, Gaze2Act omits the heatmap and keeps the contour prompt. This avoids emphasizing a fixation point that is no longer supported by the visible object region, which can occur when the dexterous hand occludes the target during contact. The gaze token follows the same validity check by using $p_t$ when valid and the mask center otherwise.

\subsection{Training and Implementation Details}
\label{app:training_details_subsec}

\paragraph{Training setup.}
Each task group is trained for 20{,}000 steps on a single NVIDIA RTX PRO 6000 GPU with 96\,GB memory and batch size 80. We use AdamW with $\beta_1 = 0.95$, $\beta_2 = 0.999$, weight decay $10^{-5}$, learning rate $1 \times 10^{-4}$, cosine decay, 5\,\% warmup, and bf16 mixed precision. The action head uses flow matching~\citep{lipman2022flow} time sampling with $\alpha = 1.5$ and $\beta = 1.0$, action horizon $H = 50$, and 4-step Euler integration at inference. The Eagle2 language model, text encoder, and DINOv3 encoders are frozen. Trainable modules include the vision encoder, DiT action head, mask control encoder, and action encoder/decoder. During training, visual prompts and gaze tokens are generated from offline ground-truth masks and fixation points; at inference, they are generated online from Meta Aria gaze.

\paragraph{Lightweight module.}
Compared with vanilla GR00T N1.5, Gaze2Act increases the parameter count from 2.414B to 2.533B, adding only 119.5M parameters (4.95\%). The extra parameters mainly come from the frozen 29M-parameter DINOv3 object encoder and lightweight gaze-conditioning layers.
\subsection{Evaluation Protocol}

Each task condition is evaluated over 50 trials. Objects are rearranged within the workspace between trials to prevent memorization of fixed spatial layouts. Intent Accuracy measures whether the system reaches toward the user-intended object or part. Task Success measures whether the full manipulation objective is completed. For part-level tasks, detector-based baselines do not produce part-specific conditions, so their part-level Intent Accuracy is not applicable; their Task Success is still reported when the object-level condition allows the policy to attempt the task. Dynamic Intent Steering is evaluated separately because the target changes during execution.

\section{Task Descriptions}
\label{app:tasks}

This section describes in detail the design motivation and specific setup for the seven task categories in Section~\ref{sec:setup}.

(i) Ambiguous Obj. (object-level disambiguation): each scene contains two or three visually similar objects from the same category, such as cups, bread, or fruits. The language instruction does not provide a unique semantic handle for the target, so the task tests whether gaze can specify the intended instance among similar distractors.

(ii) Unseen Obj. (object-level generalization): this setting uses the same training demonstrations as Ambiguous Obj., but replaces the evaluation objects with unseen instances while preserving similar-object distractors. Unseen Cup uses new cup appearances, such as an orange-patterned yellow cup; Unseen Bread uses pineapple buns; and Unseen Fruits uses unseen fruit-like objects such as tomatoes and peaches. The task tests whether gaze grounding generalizes beyond the training appearances.

(iii) Transparent Obj. (object-level perceptual challenge): transparent cups and bottles lack clear visual boundaries in RGB images and are difficult for standard detectors to segment reliably, while the explicit contour of the visual prompt can reconstruct foreground cues in the observation image.

(iv) Compositional (combinatorial spatial specification): both the manipulation object and the placement target must be specified simultaneously, and a single language instruction struggles to encode two independent spatial targets without ambiguity.

(v) Subpart Grasp (part-level localization): under the same language instruction, gaze fixation on different parts leads to different grasp locations.

(vi) Part-cond. Act. (part-level action): interacting with different regions of the same cup corresponds to different manipulation modes, such as handover or pouring; object-level masks cannot encode this part-level intent, requiring the gaze heatmap to provide sub-object resolution.

\paragraph{Dynamic intent steering.}
The operator shifts gaze from one object to another during execution while the language instruction remains unchanged. This task is more difficult than static selection because target switching requires online re-grounding under large ego-exo viewpoint changes, perspective-induced scale differences, and visually similar nearby cups, followed by redirection of an already initiated motion. It evaluates whether gaze can serve as a real-time intent update signal, rather than only a static target selector.

\begin{table}[t]
\caption{Training demonstrations and language conditioning for each task group.}
\label{tab:data}
\centering
\small
\resizebox{\textwidth}{!}{
\begin{tabular}{l l c p{4.1cm} p{4.1cm}}
\toprule
Task group & Task variants & Demos & Baseline instruction & Gaze2Act instruction \\
\midrule
Ambiguous / Unseen Obj. & Cup & 56 & handover the blue/pink patterned cup to the human. & handover the outlined object to the human. \\
Ambiguous / Unseen Obj. & Bread & 50 & handover the croissant/cream bun to the human. & handover the outlined object to the human. \\
Ambiguous / Unseen Obj. & Fruits & 50 & handover the dark red/light red fruit to the human. & handover the outlined object to the human. \\
\midrule
Transparent Obj. & Bottle / cup & 50 & handover the transparent bottle/transparent cup to the human. & handover the outlined object to the human. \\
\midrule
Compositional & Pick bread place bowl & 89 & pick the cream bun/croissant and place it in the blue or white bowl. & pick the outlined bread and place it in the outlined bowl. \\
Compositional & Pick paper ball place bin & 96 & pick the white/yellow paper ball and place it in the green/yellow bin. & pick the outlined paper ball and place it in the outlined bin. \\
\midrule
Subpart Grasp & Hammer handle / head / neck & 105 & pick up the hammer by its handle/head/neck. & Handover highlight part of the outlined object. \\
Part-cond.\ Act. & Cup handover / pouring & 100 & handover the cup. / pour water with the cup. & use the outlined object at the highlighted region. \\
\midrule
Dynamic Intent Steering & Target switch & 57 & handover the specified cup to the human. & handover the outlined object to the human. \\
\bottomrule
\end{tabular}
}
\end{table}

\section{Training Demonstrations}
\label{app:data}

Table~\ref{tab:data} summarizes the robot demonstration trajectories used for training and the language conditioning used by different methods. All methods use the same trajectories within each task group, but the trajectories are converted into different conditioning formats according to each method. Language-conditioned baselines use the most specific instruction available in the scene, while Gaze2Act uses category-level instructions and receives the spatial referent from gaze. Specifically, our data collection is tailored to various task scenarios and distractor configurations to ensure comprehensive coverage. For example, in the bread-picking task group, we collected 25 trajectories for the croissant and 25 for the cream bun, resulting in a total of 50 trajectories for that category.

Ambiguous Obj. and Unseen Obj. share the same training demonstrations. The Unseen Obj. setting differs only at evaluation time, where the test objects are replaced by unseen instances. Unseen Cup uses cups with new appearances, such as an orange-patterned yellow cup. Unseen Bread uses pineapple buns. Unseen Fruits uses unseen fruit-like objects such as tomatoes and peaches. Other evaluation settings follow the task variants in Table~\ref{tab:data}.

For Gaze2Act, target fixations are annotated offline on demonstration videos and converted into robot-view masks and fixation points using the grounding pipeline in Section~\ref{sec:grounding}. These annotations construct the visual prompts and gaze tokens during training. At inference time, the same mask-point representation is generated online from Meta Aria gaze, so eye-tracking hardware is required only for interactive deployment, not for collecting training demonstrations.

\end{document}